# An Evaluation of DUSt3R/MASt3R/VGGT 3D Reconstruction on Photogrammetric Aerial Blocks

Xinyi Wu [a,b,c], Steven Landgraf [d], Markus Ulrich d and Rongjun Qin [a,b,c,e]*

[a]Geospatial Data Analytics Lab, The Ohio State University, Columbus, OH, USA.

[b]Department of Civil, Environmental and Geodetic Engineering, The Ohio State University, Columbus, OH, USA.

[c]Department of Electrical and Computer Engineering, The Ohio State University, Columbus, OH, USA.

[d]Institute of Photogrammetry and Remote Sensing (IPF), Karlsruhe Institute of Technology, Karlsruhe, Germany.

[e]Translational Data Analytics Institute, The Ohio State University, Columbus, OH, USA.

**ABSTRACT**

State-of-the-art 3D computer vision algorithms continue to improve on sparse, unordered image sets. Recently developed foundational models for 3D reconstruction, such as Dense and Unconstrained Stereo 3D Reconstruction (DUSt3R), Matching and Stereo 3D Reconstruction (MASt3R), and Visual Geometry Grounded Transformer (VGGT), have attracted considerable attention due to their ability to handle very sparse image overlaps, as well as their generalization capability. In light of this contribution, evaluating DUSt3R/MASt3R/VGGT on typical aerial images is important, as these models may hold the potential to handle extremely low image overlaps, stereo occlusions, and textureless regions. For highly redundant collections, they can accelerate 3D reconstruction by using extremely sparsified image sets. Despite being tested on various computer vision benchmarks, their potential on photogrammetric aerial blocks remains unexplored. We present a comprehensive evaluation of the pre-trained DUSt3R/MASt3R/VGGT models on the aerial blocks of the UseGeo dataset for pose estimation and dense 3D reconstruction. The methods reconstruct dense point clouds from very sparse inputs (fewer than ten images, resized to a maximum dimension of 518 pixels), achieving reasonable accuracy and completeness gains up to 50% over COLMAP. VGGT further shows superior computational efficiency, scalability, and more reliable camera pose estimation. However, all three show limitations on high-resolution imagery and large image sets, with the camera pose estimation reliability significantly declining as the number of images and the geometric complexity of the scene increase. These findings indicate that while transformer-based method cannot replace traditional SfM and MVS methods entirely, they hold potential as complementary approaches, especially in challenging, low-resolution, and extremely sparse scenarios.

**CONTACT:** Rongjun Qin
Correspondence: Rongjun Qin qin.324@osu.edu

## 1 Introduction

Image-based 3D reconstruction and mapping enable diverse applications, such as environmental change monitoring (Jassoom and Abdoon 2024; Purkis and Klemas 2011), disaster response (Vetrivel et al. 2018; Gonsoroski et al. 2023; Pi et al. 2020), virtual and augmented reality (Noh et al. 2009), mobile 3D reconstruction (Bianco et al. 2018), computer graphics (Izadi et al. 2011), video games (Brown and Hamilton 2016), and common geomatics tasks (Hamal and Ulvi 2024; Ruan et al. 2023; Albanwan et al. 2024). Photogrammetric 3D reconstruction is a core technique that leverages rigorous perspective geometry to generate dense, accurate environmental models, often from aerial imagery. Typically, photogrammetric imagery is collected with generous overlaps (60-80%) and high redundancy, ensuring sufficient observations for robust bundle adjustment and dense image matching. However, this approach can require lengthy processing times, which limits its applicability for time-sensitive applications such as real-time mapping and disaster response planning. In addition, traditional photogrammetric methods are vulnerable when image overlap is limited, which can lead to suboptimal camera networks, occlusions, and large parallax that challenge dense surface reconstruction.

In recent years, learning-based approaches for 3D reconstruction have gained significant attention. These methods estimate an object's or scene's 3D structure end-to-end, removing the need for traditional multi-stage steps such as keypoint detection and matching. Because these models embed contextual information in pre-trained weights, they can produce high-quality reconstructions from only a handful of views (Liu et al. 2023; Pan et al. 2019; Sinha et al. 2016), and, in some cases, even from a single image (Samavati and Soryani 2023). Such methods are particularly effective for highly sparse and low-overlap datasets, offering advantages such as rapid processing. With their growing prominence, there is increasing interest in evaluating their performance in aerial photogrammetry.

The computer vision and photogrammetry communities have proposed many deep learning-based solutions for 3D reconstruction (Dame et al. 2013; Mildenhall et al. 2021; Zhan et al. 2018; Zhu et al. 2022), demonstrating different levels of performance across diverse datasets, including small indoor objects and outdoor ground-perspective scenes (Farshian et al. 2023). Among many of these pre-existing methods, DUSt3R (S. Wang et al. 2024), its sibling MASt3R (Leroy et al. 2025), and the subsequent VGGT (J. Wang et al. 2025) generalize effectively across diverse scenes. These models follow an end-to-end paradigm that predicts point clouds directly from single or stereo images, which bypasses the traditional two-step process of sparse reconstruction followed by dense reconstruction and enhances robustness to occlusions. With global motion averaging as a post-processing step, DUSt3R and MASt3R can integrate multiple images using 3D point clouds predicted from individual stereo pairs. VGGT further advances the pipeline with a feed-forward neural network that eliminates costly iterative post-optimization used by DUSt3R. As a result, VGGT may outperform DUSt3R and MASt3R in both speed and quality. Using learned priors and direct 3D registration, DUSt3R, MASt3R, and VGGT can handle individual stereo pairs and, by extension, multiple images with very low overlap and large occlusions. This suggests potential in challenging cases with only a sparse set of images, whether because the data were passively collected (for example, historical photos), resources to acquire new data are limited (for example, aerial or satellite imaging with infrequent collection), or the goal is to reach real time or near real time performance with fewer images. Despite strong results on computer vision benchmarks such as CO3Dv2 (Reizenstein et al. 2021), ETH3D (Schops et al. 2017), RealEstate10k (Zhou et al. 2018), BONN (Palazzolo et al. 2019), and the Map-free benchmark (Arnold et al. 2022), DUSt3R, MASt3R, and VGGT have not been extensively evaluated on aerial imagery. Compared to computer vision benchmarks, photogrammetric aerial images consist of rather small baselines with mostly nadir views of relatively large scenes, leading to fewer perspective variations that DUSt3R/MASt3R/VGGT typically process. Therefore, understanding their effectiveness, capabilities and accuracy potential when dealing with aerial photogrammetric images with varying density is pivotal for their practical value in the context of 3D mapping. Specifically, AerialMegaDepth (Vuong et al. 2025), designed for air-to-ground matching, proposes a scalable framework for generating pseudosynthetic data that simulates a wide range of aerial viewpoints. This framework was trained on several state-of-the-art algorithms and has demonstrated superior performance compared to the original version of DUSt3R. However, to ensure a fair comparison, this enhanced version was not included in our evaluation.

In this work, we present the first comprehensive assessment of DUSt3R, MASt3R, and VGGT for 3D reconstruction on aerial photogrammetric image blocks. We feature their strengths and limitations for pose estimation and dense point cloud generation under varying image network configurations. We use the UseGeo dataset (Nex et al. 2024) and compare these methods with COLMAP (Schönberger et al. 2016; Schonberger and Frahm 2016), a general-purpose Structure-from-Motion (SfM) and Multi-View Stereo (MVS) pipeline. **Figure 1** illustrates an example where we evaluate both dense point-cloud quality and estimated camera poses.

Our results show that classic methods remain the most effective choice for standard photogrammetric overlap rates between 60% and 80%. In contrast, VGGT serves as a valuable supplement in extremely sparse image scenarios where traditional methods fail, and it offers better scalability, efficiency, and pose estimation than DUSt3R and MASt3R.

The remainder of the paper is organized as follows. Section 2 reviews related work, covering state-of-the-art 3D modeling solutions and evaluation methods. Section 3 details the dataset configuration and evaluation metrics. Section 4 presents experimental results, analysis, and a brief case study that shows the potential of learning-based methods. Finally, Section 5 concludes our study.

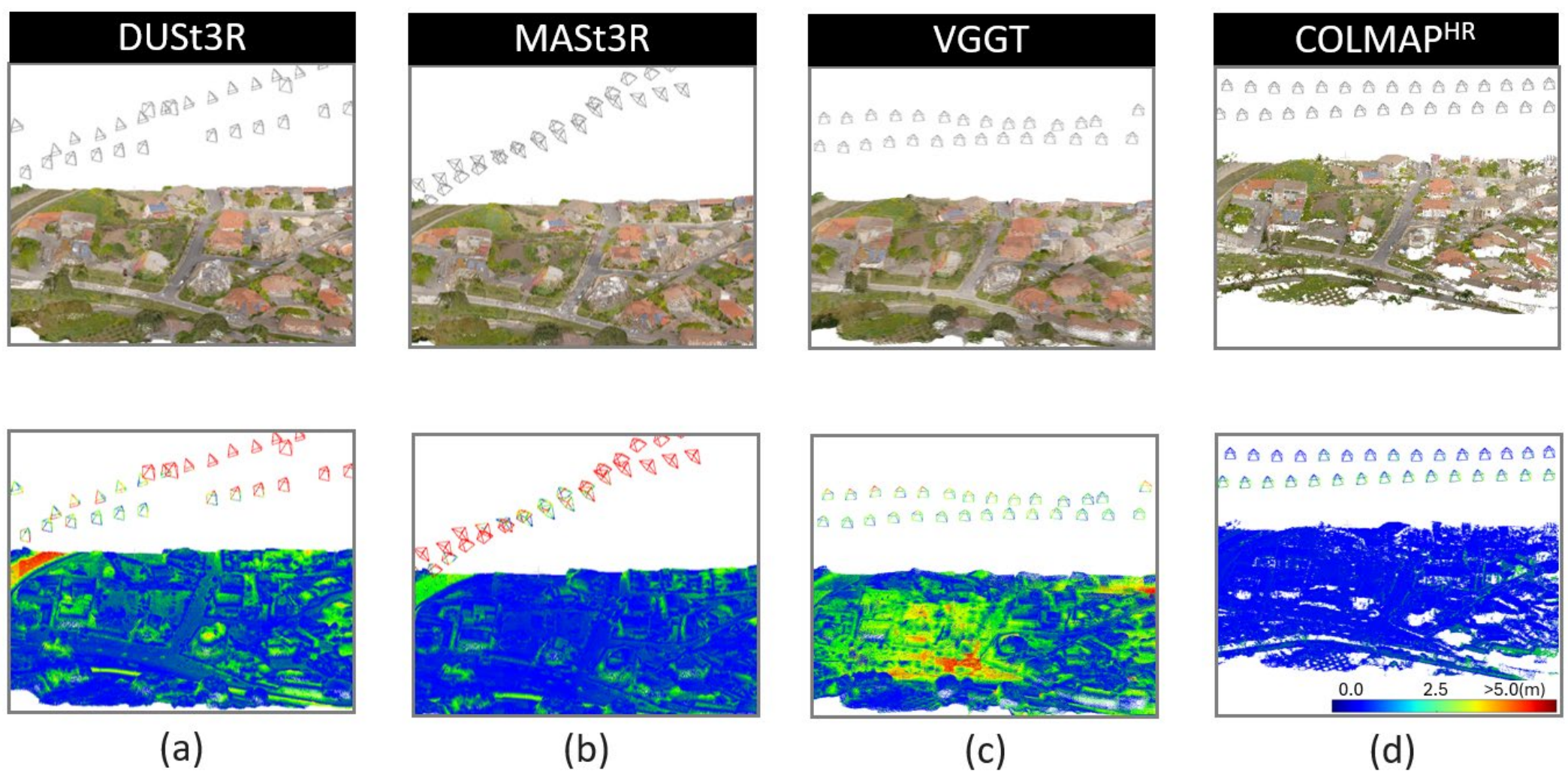

Figure 1. Results from DUSt3R (a), MASt3R (b), VGGT (c), and COLMAP$^{HR}$ (d), where COLMAP$^{HR}$ denotes COLMAP results obtained from high-resolution inputs. The top row presents the dense point cloud and the estimated camera poses (represented in gray), while the bottom row displays the error map, comparing the results to ground truth LiDAR data. Camera poses are color-coded based on their distance from the ground truth.

## 2 Related Work

Image-based 3D reconstruction has advanced rapidly in photogrammetry and computer vision. In this section, we review related work in 3D reconstruction, comparing traditional Structure-from-Motion (SfM) and Multi-View Stereo (MVS) with more recent learning-based approaches. We also examine existing evaluation studies and highlight their limitations.

**SfM and MVS.** Camera orientation and dense image matching have been widely studied, leading to the development of various algorithms and open-source tools. SfM (Crandall et al. 2011; Hartley 2003; Schonberger and Frahm 2016) processes unordered images to recover camera parameters and produce a sparse point cloud. It uses correspondences between overlapping images to compute intrinsic and extrinsic parameters (Koutsoudis et al. 2014), followed by bundle adjustment to refine camera poses (Snavely et al. 2006). Bundler by Snavely et al. (Snavely et al. 2006) is one of the earliest open-source systems for image-based 3D reconstruction and point-cloud generation. It addresses the SfM problem by estimating camera parameters. Building on this foundation, later works extended these techniques to large-scale scene reconstruction (Agarwal et al. 2011). Further, Patch-based Multi-View Stereo (PMVS), introduced by Furukawa and Ponce (Furukawa and Ponce 2010, 20), performs for dense image matching to produce detailed reconstructions. More broadly, MVS reconstructs dense point clouds from a set of images, and the final 3D model is obtained fusing per-view depth maps into a single coherent representation. These tools have been widely adopted by researchers and practitioners (Furukawa et al. 2015). Numerous frameworks and libraries have since been released, extending these techniques. Examples include the Multi-View Environment (MVE) (Furukawa et al. 2015), an end-to-end pipeline for image-based geometry reconstruction, and Open Multiple View Geometry (OpenMVG) by Moulon et al.

(Moulon et al. 2017), a library tailored to the multiple-view geometry community. More recently, full 3D reconstruction pipelines such as COLMAP and OpenMVS (Cernea 2020) provide comprehensive solutions for a broad audience. In parallel, advances in deep learning for computer vision and photogrammetry have increased the prominence of learning-based approaches (Hartmann et al. 2017; Kerbl et al. 2023; C. Wang et al. 2024), particularly in areas such as self-supervised methods for single-image depth estimation (Knöbelreiter et al. 2018; Madhuanand et al. 2021).

**Direct RGB-to-3D.** Unconstrained dense 3D reconstruction from multiple RGB images remains a long-standing research problem in 3D modeling (Charles et al. 2017; Dame et al. 2013; Mildenhall et al. 2021). In recent years, neural network-based methods that predict depth from a single image or a very small number of images have gained significant attention. These approaches, used not only for matching (Ji et al. 2019), address many limitations of two-view and multi-view stereo depth estimation. Notably, they eliminate the sequential dependency of the SfM pipeline, which tends to accumulate errors and noise at each processing stage. Some methods use neural networks to learn robust geometric class-level priors or diffusion models (Liu et al. 2023). However, these approaches are primarily designed for object-centric reconstruction rather than large-scale scene reconstruction. Another line of research focuses on general scene reconstruction by using monocular depth estimation neural networks trained on large datasets. These methods can produce pixel-aligned 3D point clouds (Ranftl et al. 2021; Wiles et al. 2020; Yin et al. 2021), although depth quality can lack fidelity because of missing scale or out-of-distribution prediction. To address this limitation, multi-view neural networks for direct 3D reconstruction have been introduced, which enable end-to-end training and resolving scale ambiguity (Ummenhofer et al. 2017). More recently, DUSt3R has emerged as a notable advance, eliminating the need for ground truth camera intrinsics as input. This approach can directly generate point maps and global camera poses rather than relying on depth maps and relative poses. The promising results of DUSt3R and its sibling MASt3R have driven further progress, inspiring the development of more sophisticated methods such as VGGT (Visual Geometry Grounded Transformer) (J. Wang et al. 2025). VGGT is a feed-forward neural network built on a standard large transformer(Vaswani et al. 2017). It removes pairwise point cloud generation and can process more than two images simultaneously, enabling direct production of point clouds without post-processing to fuse pairwise reconstructions. This design can yield more consistent point cloud results.

As interest grows, new models are appearing rapidly. Fast3R (Yang et al. 2025) extends the DUSt3R family to a single forward pass designed for large N inputs, improving throughput from a handful of views to hundreds or more. Along the VGGT line, FastVGGT (Shen et al. 2025) and Faster VGGT (C.-S. B. Wang et al. 2025) identify global attention as the main bottleneck: the former uses token merging and the latter uses optimized block sparse attention to accelerate inference while keeping quality comparable. These methods are promising for aerial applications because they are efficient and can scale to hundreds or even thousands of images. In quick tests against the learning-based methods used in this paper, the newest models showed similar or slightly better point cloud and pose quality. Metrics include point cloud accuracy and completeness as well as pose center and orientation errors. These findings do not change our conclusions. Given the fast pace of the field and our scope, we proceed without adding these models and instead cite them because of their recent release.

**Surveys, Reviews, and Evaluation.** With the rise of open-source 3D reconstruction solutions, evaluating these pipelines has become common in the research community. Reviews have analyzed methods, datasets, scenarios, and photogrammetric metrics (Alidoost and Arefi 2017; Georgopoulos et al. 2016; Pepe et al. 2022). Moreover, Remondino et al. (Remondino et al. 2017) documented the development of diverse MVS algorithms for reconstructing different scenes. Stathopoulou et al. (Stathopoulou et al. 2019) examined widely used open-source image-based 3D reconstruction pipelines, while Jarahizadeh and Salehi (Jarahizadeh and Salehi 2024) presented a recent evaluation of popular photogrammetry software. However, these efforts are limited to traditional MVS solutions. Learning-based methods have gained attention, and new evaluation practices have appeared because these approaches have the potential to surpass traditional methods in multiple domains. Unlike conventional techniques, they support end-to-end training, which removes the need for manually designed multi-stage processes. Several studies have surveyed key challenges, network architectures, and evaluation methodologies in 3D reconstruction (Fahim et al. 2021; Fu et al. 2021). However, their scope is limited to single-image 3D object reconstruction. Han et al. (Han et al. 2021) extend the scope by covering both single- and multi-image , but they do not include research published after 2019 and thus miss recent advances. Additionally, Samavati and Soryani (Samavati and Soryani 2023) take a broader perspective by exploring studies where 3D reconstruction serves as a downstream task for various objectives. Their survey mentions DUSt3R but does not provide experimental data to support its performance.

The rapid progress of the field calls for regular reassessment of recent research. Evaluating new methods on updated benchmark datasets is essential to keep pace with ongoing advances.

## 3 Material Preparation and Experiment Setup

This section presents the benchmark dataset and our data preparation workflow, then outlines the evaluated approaches for 3D reconstruction. Finally, we define the metrics used to assess dense point clouds and camera poses.

### *3.1 Dataset Configuration*

We use the UseGeo dataset (Nex et al. 2024), which includes images and LiDAR collected at the same time across diverse urban and peri-urban areas. The UseGeo dataset is intended for rigorous benchmarking in the context of photogrammetry applications. A total of 829 high-resolution images were captured at an average altitude of 80 m during three flights that cover three distinct areas, which we refer to as Dataset-1, Dataset-2, and Dataset-3. Each dataset contains eight flight strips, with typical image overlap of 60–80%. LiDAR was acquired simultaneously at about 51 points per square meter, which corresponds to a Ground Sample Distance (GSD) of approximately 2 cm. Following image and LiDAR acquisition, the hybrid adjustment (Glira et al. 2019) method was employed to jointly refine the orientations of the LiDAR and camera, optimizing image alignment, camera calibration, and distortion correction. The adjusted LiDAR and camera data serve as ground truth (GT) for both point cloud accuracy and camera pose estimation. In the UseGeo dataset, the mean cloud-to-cloud (C2C) residual error between LiDAR and photogrammetric point clouds is 6.7–8.8 cm, which indicates strong internal alignment. Additional preprocessing details are provided in dataset paper (Nex et al. 2024). The dataset challenges learning-based methods because the limited number of images and their somewhat homogeneous (nadir) perspectives. Although overlap is sufficient for classic SfM, it can be relatively small for self-supervised methods that rely on joint depth and relative motion estimation (Hermann et al. 2024).

To study performance across coverage levels, we evaluated subsets of 1, 2, 5, 10, and 38 images from Dataset-1, Dataset-2, and Dataset-3 for the main experiments. **Figure 2** top row shows examples of the selected images in Dataset-1. The 38-image subset is the largest that DUSt3R and MASt3R can handle on our hardware because these models are computationally intensive. For scalability analysis, we included an experiment with 191 images which is the maximum we can process with VGGT on our device. For the 191 image experiment, only VGGT and COLMAP were evaluated, as they are capable of handling datasets of this scale. In these experiments, images were typically acquired along one to five flight strips, with the number of strips varying according to the number of images selected and the specific area of interest. For the scalability experiment, the 191-image subset comprised the first 191 images from each dataset. The full list of the image IDs are provided in Appendix B. We selected images from different datasets with varying scene complexity; examples are shown in **Table 1**.

Furthermore, to test robustness under low-overlap photogrammetric blocks, we conducted an experiment, referred to as low-overlap reconstruction, which reduced the original overlap rate from approximately 70% to 10% with 38 images. **Figure 2** bottom row shows low overlap experiments with 70, 55, 40, 25, and 10% overlap in Dataset-1. To systematically reduce image overlap in our experiments, we primarily decreased the along-track overlap by selecting images at larger intervals along the flight path, while keeping across-track coverage largely unchanged (Torres-Sánchez et al. 2018). The areas of interest were first identified, and images capturing these regions were selected. When images were chosen sequentially along different drone flight trajectories, the overlap was around 70%. Selecting every other image (i.e., skipping one) reduced the overlap to approximately 55%. Similarly, skipping two images resulted in a 40% overlap, skipping three images led to 25%, and skipping four images reduced the overlap to about 10%. This sampling strategy enabled us to assess the sensitivity of each reconstruction method to reduce along-track redundancy, which is relevant for scenarios with limited acquisition resources or the need for faster processing. Naturally, at higher overlap rates, the selected images were concentrated in a smaller region, whereas at lower overlap rates, the images were more spatially distributed, potentially covering a larger area.

DUSt3R and MASt3R use transformer architecture and, on mainstream GPUs as of 2025, are limited to images with a maximum lateral dimension of 512 pixels. VGGT requires input images with a maximal dimension of 518 pixels. Consequently, we rescaled all images to 512 pixels to fit these limits while preserving aspect ratios. For COLMAP, we report results on both the rescaled images, where the largest dimension is 512 pixels for fair comparison, and the original-resolution images to assess real-world performance. Here, $\text{COLMAP}^{\text{HR}}$ refers to high-resolution inputs and $\text{COLMAP}^{\text{LR}}$ to the low-resolution setting.

Beyond the benchmark datasets, we include a practical case study on a self-collected aerial dataset to show how learning-based solutions reduces coverage gaps. The data were collected on The Ohio State University (OSU) campus with a high-quality Unmanned Aerial Vehicle (UAV). The dataset contains 1,190 images with approximately 80% overlap. Each image has GPS with about 0.25 m positional accuracy. The camera is a DJI FC6310S with a 9 mm focal length and an image size of 5472 by 3648 pixels. The scene is complex, with tall and low buildings and detailed facades. Fixed pattern flights leave sparse views near vertical surfaces, which often produce incomplete meshes This dataset is a good example of when learning-based methods help. The case study uses the same pipeline and is reported in Section 4.5.

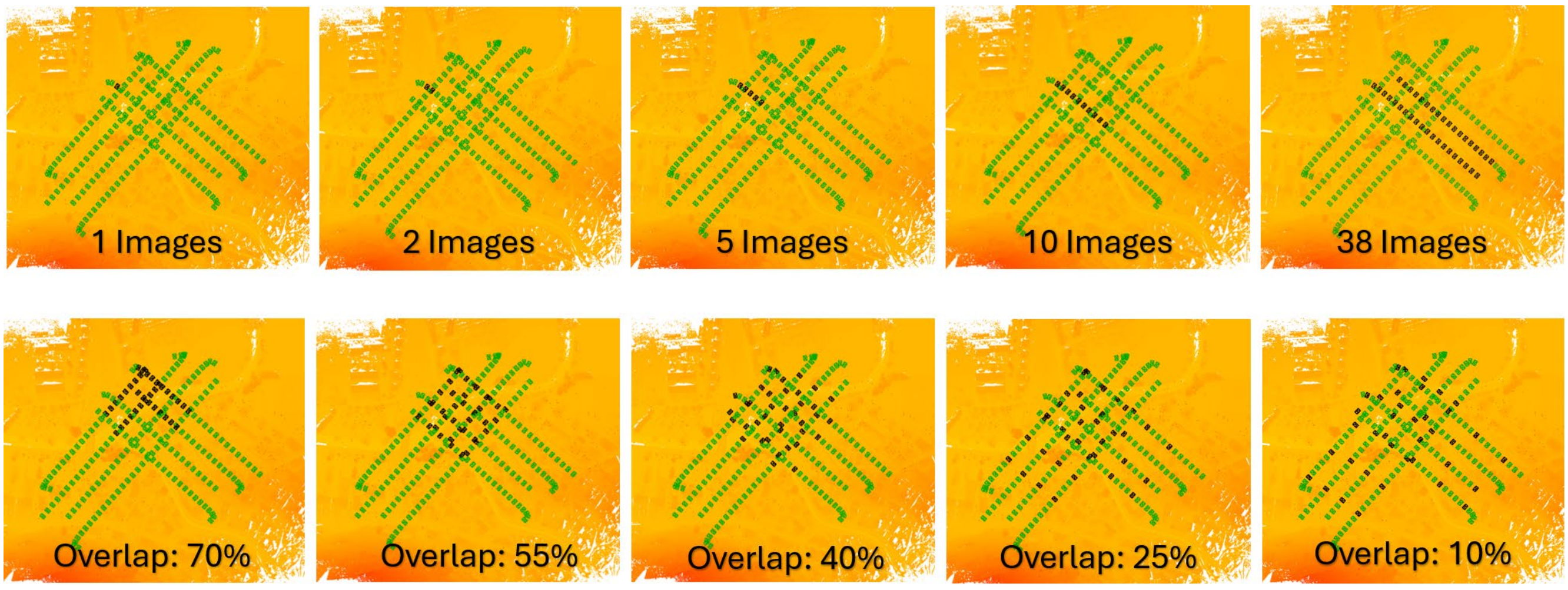

**Figure 2.** Spatial distribution of the selected cameras for Dataset-1. Top row experiments with 1, 2, 5, 10, and 38 images. Bottom row low overlap experiments with 70, 55, 40, 25, and 10% overlap. Each square marks a camera position. Green points are all ground truth poses. Black points are the selected poses. The background is the ground truth LiDAR point cloud color coded by elevation.

### *3.2 Evaluated Methods*

DUSt3R is a transformer-based method that works without prior knowledge of camera calibration or viewpoint poses. It treats pairwise reconstruction as a regression from image to point maps, which bypasses the strict constraints of traditional projective camera models (S. Wang et al. 2024). MASt3R extends DUSt3R by adding a second network head to generate dense local features that are trained with a newly introduced matching loss. Although MASt3R shows strong overall performance across a range of matching tasks, including those in which it outperforms DUSt3R, it is restricted to the binocular case and lacks an implementation for multiple images (Leroy et al. 2025). To compare multi-image reconstructions, we applied the global alignment strategy from the DUSt3R paper to MASt3R's pairwise outputs, aligning point maps into a single reference frame. Specifically, AerialMegaDepth provides a scalable framework for generating pseudosynthetic data that simulates diverse aerial viewpoints. State-of-the-art algorithms, such as DUSt3R finetuned on this dataset, have shown stronger performance than the original version of DUSt3R. However, this enhanced version was not included in our evaluation to ensure a fair comparison.

VGGT introduces a feed-forward neural network that performs 3D reconstruction directly from as few as one view and up to hundreds of views, which removes the need for post-processing geometry optimization. This approach offers more consistent point clouds, reduces the computational cost of iterative optimization, and has the potential to outperform DUSt3R and MASt3R by a substantial margin.

COLMAP (Schönberger et al. 2016; Schonberger and Frahm 2016) is a general-purpose SfM and MVS pipeline. It uses Scale-Invariant Feature Transform (SIFT) (Lowe 2004), for feature extraction and matching, followed by geometric validation, incremental SfM, and bundle adjustment to refine camera and point estimates (Stathopoulou et al. 2019). Further, a probabilistic patch-based stereo framework was used for MVS reconstruction. Except for setting the minimum number of reconstructed images for an accepted model to two, all COLMAP parameters were left at default settings to ensure consistency and provide a baseline for comparison. We used the defaults because we did not find existing literature reporting any aerial-specific settings. The defaults also are a well-known, general-purpose baseline that supports transparency and comparability. To address concern about tuning, we include a small preliminary experiment in Appendix A. We vary the matching ratio and the bundle adjustment iteration budget and find that these changes do not materially affect the conclusions when compared with the learning-based methods.

In this study, we evaluated DUSt3R, MASt3R, VGGT, and COLMAP on our datasets in terms of reconstruction accuracy and robustness. The main settings are recorded in **Table 2**. Meanwhile, pre-trained models were used. DUSt3R employed a model trained on the rescaled images in which the largest dimension is 512 pixels, with the dense prediction transformer (DPT) head (Ranftl et al. 2021), while MASt3R utilized a model trained on similar rescaled images with a mixed multi-layer perceptron (MLP) and DPT architecture (termed CatMLP+DPT). This architecture combines an MLP and a DPT head, where the MLP outputs 3D points and local features. Both heads receive input from a concatenation of the encoder and decoder outputs. VGGT rescales input images to a width of 518 pixels while maintaining the aspect ratio. It utilizes a unified architecture with a ViT-Large transformer encoder and no separate decoder, employing multiple task-specific heads for outputs such as camera parameters, depth, and point clouds. We performed end-to-end training with a multi-task loss and used mixed precision to reduce runtime and memory use. DUSt3R and MASt3R ran with a batch size of 1 and a maximum input dimension of 512 pixels, and VGGT ran with a batch size of 1 and an input width of 518 pixels.

We align the reconstructed point clouds and camera poses to the ground truth model independently. Point cloud alignment involved an initial manual alignment, followed by refinement using the iterative closest point (ICP) algorithm (Besl and McKay 1992) implemented in CloudCompare (Girardeau-Montaut and others 2016). For camera poses, we align the estimated positions to the ground truth by solving two rigid transformations in sequence that model scale, rotation, and translation with a least-squares approach. The transformations were first applied to the camera centers, followed by the orientations, and then combined to produce the final alignment.

Table 1 An overview of the scenarios and datasets used in this evaluation, including example

photogrammetric point clouds generated for the test areas and the ground truth (GT). The GT point cloud is color-coded by height, with GT camera poses overlaid. Scale bars are included in the GT visualizations.

| Type | # Images | # GT Points | Example point clouds | GT |
|---|---|---|---|---|
| Dataset-1 | 224 | 105.9 M | | |
| Dataset-2 | 327 | 146.3M | | |
| Dataset-3 | 277 | 140.6M | | |

Table 2 Overview of key modules in traditional (COLMAP) and learning-based (DUSt3R/MASt3R/VGGT) 3D reconstruction pipelines. DLT: Direct Linear Transformation.

**Traditional Methods**

| | Feature Extraction | Feature Matching | Geometric Verification | Image Registration | Triangulation | Robust Estimation | Dense point cloud generation |
|---|---|---|---|---|---|---|---|
| **COLMAP** | SIFT (Lowe 2004) | Exhaustive search | 7-Point F-matrix (Hartley and Zisserman 2003) | P3P (Gao et al. 2003) | Sampling-based DLT | RANSAC | Patch-based stereo (Schönberger et al. 2016) |

**Learning-based Methods**

| | Encoder | Decoder | Heads | Network Loss |
|---|---|---|---|---|
| **DUSt3R/ MASt3R** | ViT-Large (Dosovitskiy et al. 2020) | ViT-Base (Dosovitskiy et al. 2020) | DPT (Ranftl et al. 2021) / CatMLP+DPT | Simple regression loss |
| **VGGT** | ViT-Large (Dosovitskiy et al. 2020) | - | Task-specific heads | Multi-task loss |

### *3.3 Evaluation on Dense Point Clouds Generation*

**Accuracy.** Accuracy is measured using the quadratic height function in CloudCompare, which computes the vertical distance between each estimated point and the corresponding reference surface derived from the ground truth point cloud. This method provides a more reliable accuracy assessment by considering local surface variations rather than simple point-to-point Euclidean distances. The mean accuracy represents the average vertical deviation between the reconstructed point cloud and the ground truth

LiDAR data. We follow existing works (Ahmad Fuad et al. 2018; Xu et al. 2023) and use the mean C2C distance, $\sigma_{\text{MEAN}}$, as shown in Equation (1).

$$\sigma_{\text{MEAN}} = Mean\left(D_{point_to_local_surface}\right) \tag{1}$$

**Completeness.** Completeness is measured by reversing the process: the vertical distance between each ground truth point and the corresponding reference surface derived from the estimated point cloud is calculated, with an empirical threshold of 1 m applied. Completeness is defined as the ratio of ground truth points within this threshold ($N_{within}$) to the total number of ground truth points ($N_{GT}$), where $N_{within}$ is the number of ground truth points within the threshold, and $N_{GT}$ is the total number of ground truth points.

$$N_{within} = \sum_{j=1}^{N_{GT}} \delta\left(d\left(\boldsymbol{p}_{j,GT}, P_E\right) \leq \tau\right) \tag{2}$$

Where $d\left(\boldsymbol{p}_{j,GT}, P_E\right)$ is the vertical distance from the ground truth point $\boldsymbol{p}_{j,GT}$ to the corresponding reference surface derived from the estimated point cloud $P_E$. Here, $\tau$ is the threshold (e.g., 1 m); $\delta(\cdot)$ is an indicator function that equals 1 if the condition inside is true, and 0 otherwise. The evaluation employs both accuracy and completeness to provide a comprehensive analysis of the results.

### *3.4 Evaluation on Camera Poses Estimation*

The pose of each camera is compared against its corresponding ground truth, evaluating both position and orientation.

#### *3.4.1 Evaluation of Camera Position/Translation*

The camera position is assessed by calculating the Euclidean distance between the reconstructed position and the ground truth position, as shown below:

$$\Delta C = \left\|\boldsymbol{C_{pred}} - \boldsymbol{C_{gt}}\right\| \tag{3}$$

where $\Delta C$ is the camera center difference (in meters), $C_{pred}$ is the predicted camera center, $C_{gt}$ is the ground truth camera center, and $\| \cdot \|$ denotes the Euclidean norm (distance).

#### *3.4.2 Evaluation of Camera Rotation/Orientation.*

Orientation differences are assessed by determining the angle of the rotation required to align the reconstructed camera's orientation with the ground truth (Bianco et al. 2018; Xu et al. 2024). We represent orientations with unit quaternions and compute the error from the relative quaternion. The relative quaternion is calculated as follows:

$$\boldsymbol{q_R} = \boldsymbol{q_E}^{-1}\boldsymbol{q_{GT}} \tag{4}$$

Here, $\mathbf{q}_{\boldsymbol{R}}$ represents the quaternion describing the rotational transformation needed to align the estimated camera orientation ($\mathbf{q}_{\boldsymbol{E}}$) with the ground truth orientation ($\mathbf{q}_{\boldsymbol{GT}}$), where $\mathbf{q}_{\boldsymbol{E}}^{-\mathbf{1}}$ denotes the inverse of the estimated orientation. The orientation error of camera poses is measured in terms of angle difference ($\alpha$), and can be computed from the scalar part $w$ of the quaternion, as shown in Equation (5).

$$\alpha = \cos^{-1}(\boldsymbol{q}_{Rw}) \tag{5}$$

## 4 Experiment Results

First, we assess the reconstructed point clouds, focusing on accuracy and completeness as key metrics, as shown in Section 4.1. Next, we compare methods by camera-center differences and camera-angle distances, as shown in Section 4.2. The scalability study on 191 images using VGGT and COLMAP appears in Section 4.3, and Section 4.4 reports runtime and computational resources. Finally, Section 4.5 reviews the practical implications of learning-based reconstruction for aerial data.

We use COLMAP$^{HR}$ for results from high-resolution inputs and COLMAP$^{LR}$ for results from low-resolution inputs. COLMAP refers to the method family regardless of resolution. All experiments were conducted on a system running Ubuntu 22.04.5 LTS, equipped with an AMD Ryzen Threadripper PRO 5955WX CPU (16 cores, 1.8–4.0 GHz), 512 GB RAM, and an NVIDIA RTX 6000 Ada Generation GPU (52 GB VRAM).

### *4.1 Accuracy of Dense Point Clouds*

As **Figure 3** illustrates, for the single-image case, DUSt3R, MASt3R, and VGGT reconstruct dense urban point clouds, whereas COLMAP fails because viewing angles are insufficient for triangulation. However, the reconstructed models still have flaws, exhibiting holes around buildings and failures on small towers, likely due to limited model understanding of tall structures in top-down views and insufficient resolution. Similarly, when using two images with a large viewpoint difference, COLMAP often fails or produces low-quality models with sparse points, achieving an accuracy of up to 2.3 m. In contrast, DUSt3R, MASt3R, and VGGT produce reasonable point clouds, with MASt3R and VGGT showing similar performance and generally outperforming the others. These methods achieve higher accuracy (up to 0.4 m) and greater completeness (an increase of +10%), as shown in **Table 3**.

MASt3R and VGGT outperform COLMAP in completeness in 87% of instances, achieving up to an additional 19% completeness in most scenarios. This is due to their ability to generate more points without geometric constraints, unlike COLMAP, which prioritizes higher accuracy by producing fewer points. Learning-based methods such as MASt3R employ a coarse-to-fine, one-versus-all strategy for point triangulation, while VGGT directly predicts near-accurate point or depth maps. Both approaches lack epipolar constraints and multi-view consistency, which leads to denser and more efficient, but less accurate point clouds. This trade-off yields higher completeness but lower accuracy in reconstructions.

As the number of images increases, COLMAP leverages good viewing angle differences to reconstruct a model, with high-resolution input achieving significantly higher accuracy. The qualitative results for Dataset-3 using 38 images are presented in **Figure 4**. In this case, COLMAP$^{HR}$ achieves an accuracy of 0.2 m, corresponding to a 92% reduction in error compared to the other methods, which have errors around 2.0 m. One potential factor contributing to COLMAP$^{HR}$'s superior accuracy is that it processes images at higher resolutions, allowing for more precise feature extraction and matching. However, when analyzing scenarios using rescaled images with a maximum dimension of 512 pixels, COLMAP$^{LR}$'s accuracy fluctuates substantially, sometimes resulting in errors of 4 m in contrast to MASt3R's 0.4 m, and COLMAP$^{LR}$ suffers from very low completeness due to the limited number of 3D points detected.

Overall, COLMAP$^{HR}$ consistently achieves the highest accuracy when results are available and generally maintains acceptable completeness. Although its completeness is sometimes lower than that of VGGT, the difference is not substantial. Its performance is stable, especially as the number of images increases. However, MASt3R and VGGT demonstrate clear advantages in challenging scenarios with very limited images, where COLMAP often fails or cannot be applied. This suggests that, although MASt3R and VGGT are not yet a complete replacement for traditional methods in standard SfM and MVS pipelines, they can serve as a valuable supplement, particularly for improving completeness in sparse or difficult cases.

The results of the low-overlap reconstruction experiment using 38 images are presented in **Table 4**. Overall, these findings are consistent with previous observations: COLMAP achieves higher accuracy, whereas MASt3R and VGGT demonstrate comparable performance and superior completeness. Specifically, COLMAP achieves higher accuracy in 93% of cases, with accuracy up to 80% better than that of the others. In contrast, MASt3R and VGGT outperform both COLMAP variants in completeness in 80% of cases, with gains of up to +50%. Further, as the overlap decreases, the learning-based methods maintain both accuracy and completeness, exhibiting robustness in extremely low-overlap scenarios, whereas COLMAP experiences a significant performance drop in completeness (e.g., 8%), which is insufficient for practical real-world applications. Although COLMAP can generate highly accurate point clouds, its performance degrades significantly when the image overlap is reduced to 10%, which is expected since this overlap rate is outside the typical operational range for which COLMAP was designed. With limited overlap, COLMAP struggles to find correct feature matches, leading to fewer accurately matched 2D points and, consequently, fewer reconstructed 3D points. In contrast, transformer-based methods like VGGT can generate more 3D points even in low-overlap conditions, giving them a clear advantage in point cloud completeness and density.

To sum up, MASt3R and VGGT outperform COLMAP in extremely sparse views across both resolution settings, such as one or two images or approximately 10% overlap, achieving higher accuracy (up to 0.4 m) or up to +50% completeness. In contrast, COLMAP often fails in these settings or yields larger errors (up to 2.3 m) and much lower completeness (as low as 8%). Although MASt3R and VGGT demonstrate robust performance in extremely low-overlap cases, maintaining high completeness and comparable accuracy, their advantage diminishes in high-resolution photogrammetry datasets with typical overlaps (i.e., 70%). In these cases, they exhibit either similar or moderately higher completeness, with an advantage of up to 20%, while COLMAP achieves substantially greater accuracy, reducing errors by up to 9%. This comparison shows that, although transformer-based methods can provide value in special cases with limited images, COLMAP is better suited for routine photogrammetric workflows.

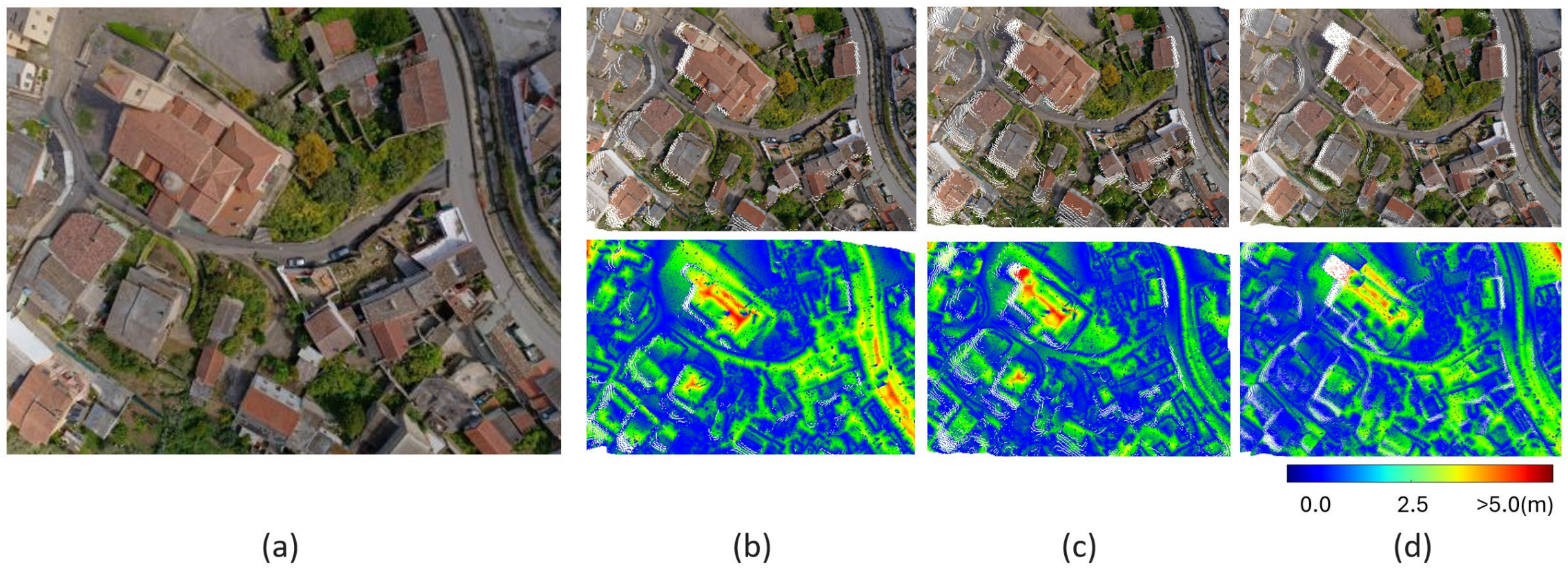

**Figure 3** Reconstruction results using a single image. (a) Input image; (b), (c), and (d) show the reconstruction results of DUSt3R, MASt3R, and VGGT, respectively. The upper row presents the dense point cloud, and the bottom row displays the error map. The color bar encodes absolute error in m: 0.0 to 2.5 m is blue through green, 2.5 to 5.0 m is green through yellow-orange, and values greater than 5.0 m are red. White denotes no data or invalid depth.

**Table 3** Quantitative evaluation of dense point cloud reconstruction across three datasets using 1, 2, 5, 10, and 38 images with different methods. “Accu.” denotes accuracy, “Comp.” denotes completeness, and ” - ” indicates no results. The best results are bolded.

| Dataset | Method | 1 Image | | 2 Images | | 5 Images | | 10 Images | | 38 Images | |
|---|---|---|---|---|---|---|---|---|---|---|---|
| | | **Accu. (m)** | **Comp. (%)** | **Accu. (m)** | **Comp. (%)** | **Accu. (m)** | **Comp. (%)** | **Accu. (m)** | **Comp. (%)** | **Accu. (m)** | **Comp. (%)** |
| Dataset-1 | DUSt3R | 0.697 | 8.780 | 0.625 | 11.81 | 0.523 | 19.52 | 0.689 | 33.56 | 0.709 | 66.52 |
| | MASt3R | **0.364** | **14.85** | 0.432 | 14.18 | 0.343 | 24.60 | 0.390 | **38.82** | 0.436 | **78.90** |
| | VGGT | 0.629 | 10.61 | **0.422** | **15.98** | 0.353 | **25.80** | 0.491 | 38.58 | 1.122 | 74.96 |
| | COLMAP$^{LR}$ | - | - | - | - | 2.625 | 2.130 | 0.535 | 6.310 | 4.161 | 17.50 |
| | COLMAP$^{HR}$ | - | - | - | - | **0.070** | 20.64 | **0.085** | 36.85 | **0.064** | 59.74 |
| Dataset-2 | DUSt3R | 2.401 | 6.230 | 0.616 | 13.16 | 0.699 | 16.17 | 0.860 | 20.51 | 1.452 | 36.42 |
| | MASt3R | 2.175 | **7.660** | 0.735 | 13.45 | 0.540 | **22.22** | 0.590 | 27.16 | 0.925 | 49.71 |
| | VGGT | **1.389** | 7.27 | **0.596** | **14.60** | 0.649 | 20.41 | 0.909 | **31.54** | 1.090 | 62.64 |
| | COLMAP$^{LR}$ | - | - | - | - | 0.590 | 12.58 | 0.859 | 20.51 | 0.325 | 61.09 |
| | COLMAP$^{HR}$ | - | - | 2.349 | 4.300 | **0.122** | 17.11 | **0.150** | 27.60 | **0.127** | **74.36** |
| Dataset-3 | DUSt3R | 1.039 | 6.720 | 0.925 | 6.980 | 0.786 | 11.92 | 0.807 | 21.82 | 2.041 | 45.78 |
| | MASt3R | 0.889 | 5.710 | **0.774** | **8.300** | 0.627 | 13.48 | 0.574 | 29.81 | 1.583 | 41.76 |
| | VGGT | **0.871** | **6.955** | 1.014 | 6.662 | 0.658 | **15.35** | 0.514 | **30.68** | 1.158 | 30.67 |
| | COLMAP$^{LR}$ | - | - | - | - | - | - | - | - | 0.288 | 55.69 |
| | COLMAP$^{HR}$ | - | - | - | - | **0.134** | 12.58 | **0.106** | 28.58 | **0.163** | **69.73** |

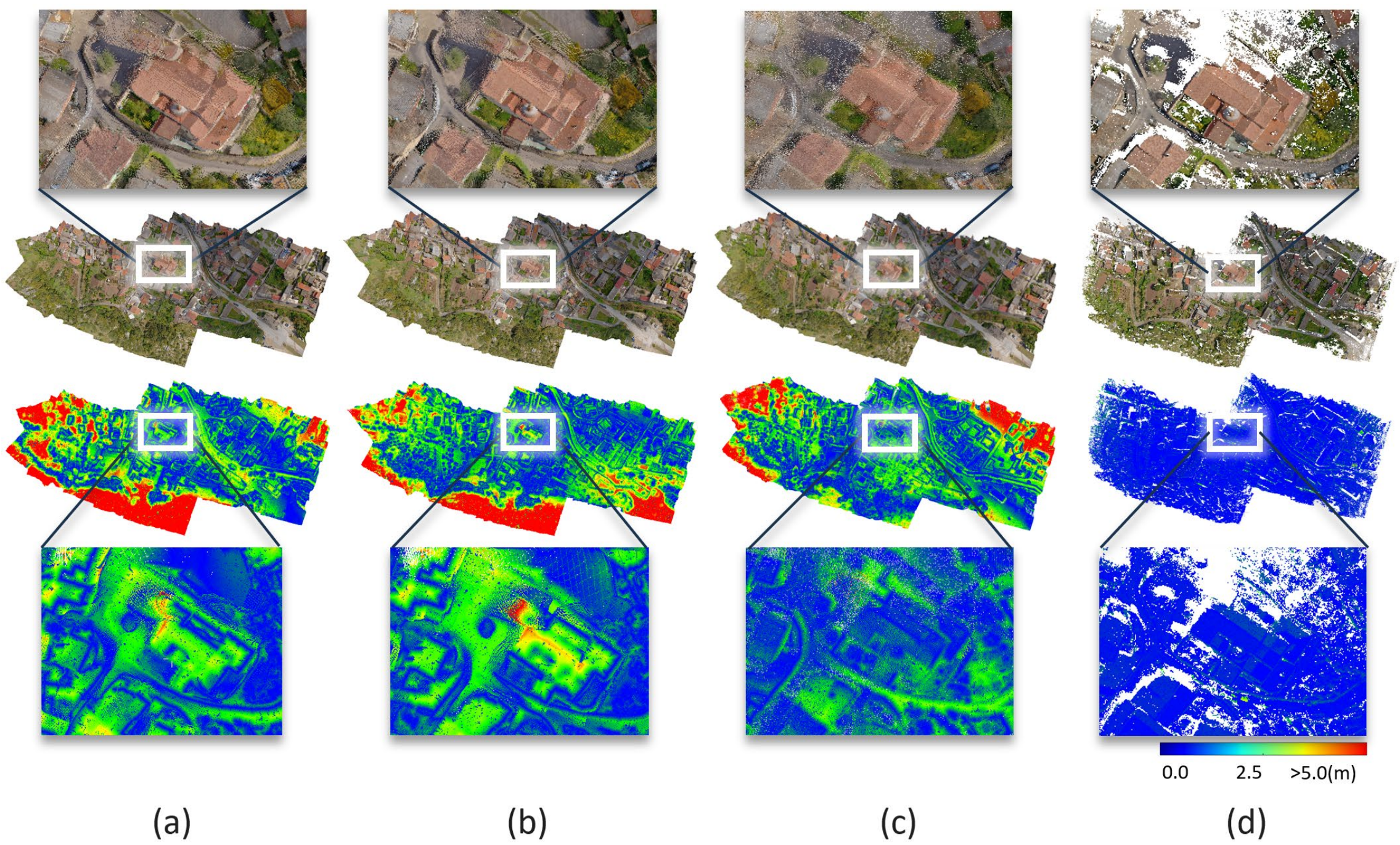

**Figure 4.** Reconstruction results using 38 images. (a), (b), (c), and (d) show the reconstruction results of DUSt3R, MASt3R, VGGT, and COLMAP$^{HR}$, respectively. The first row presents detailed views of the dense colored point clouds; the second row shows the overall dense point clouds; the third row depicts the error maps of the dense point clouds; and the bottom row highlights zoomed-in details of the error maps. The color bar encodes absolute error in m: 0.0 to 2.5 m is blue through green, 2.5 to 5.0 m is green through yellow-orange, and values greater than 5.0 m are red. White denotes no data or invalid depth.

**Table 4.** Quantitative evaluation of dense point cloud reconstruction across three datasets with different image overlaps and methods. “Accu.” denotes accuracy, “Comp.” denotes completeness, and “–” indicates no results. The best results are highlighted in bold.

| **Dataset** | **Method** | **Overlap: 70%** | | **Overlap: 55%** | | **Overlap: 40%** | | **Overlap: 25%** | | **Overlap: 10%** | |
|---|---|---|---|---|---|---|---|---|---|---|---|
| | | **Accu. (m)** | **Comp. (%)** | **Accu. (m)** | **Comp. (%)** | **Accu. (m)** | **Comp. (%)** | **Accu. (m)** | **Comp. (%)** | **Accu. (m)** | **Comp. (%)** |
| Dataset-1 | DUSt3R | 0.656 | 26.41 | 0.732 | 30.41 | 0.918 | 30.25 | 0.791 | 59.60 | 1.029 | 47.72 |
| | MASt3R | 0.433 | 30.09 | 0.533 | 34.97 | 0.707 | 36.98 | **0.612** | 59.60 | 0.919 | 52.97 |
| | VGGT | 0.542 | **31.71** | 0.507 | **37.66** | 0.862 | **39.91** | 1.087 | **58.56** | 1.589 | **58.59** |
| | COLMAP$^{LR}$ | 0.484 | 11.73 | 2.546 | 14.41 | 1.389 | 4.010 | 7.642 | 0.440 | - | - |
| | COLMAP$^{HR}$ | **0.086** | 24.10 | **0.106** | 17.81 | **0.432** | 20.02 | 0.945 | 17.23 | **0.917** | 8.12 |
| Dataset-2 | DUSt3R | 2.085 | 11.88 | 1.389 | 16.79 | 1.954 | 15.98 | 2.532 | 19.81 | 6.717 | 19.83 |
| | MASt3R | 0.898 | **28.89** | 0.924 | 23.08 | 1.682 | 24.49 | 1.432 | **30.65** | 2.518 | 26.76 |
| | VGGT | 0.708 | 21.52 | 1.412 | 25.47 | 2.331 | 14.97 | 4.413 | 21.33 | 5.810 | **35.37** |
| | COLMAP$^{LR}$ | 0.278 | 13.92 | 0.428 | 14.52 | 2.980 | 3.580 | - | - | - | - |
| | COLMAP$^{HR}$ | **0.088** | 20.43 | **0.118** | **26.16** | **0.141** | **28.24** | **0.254** | 17.90 | **0.371** | 10.71 |
| Dataset-3 | DUSt3R | 1.829 | 40.29 | 1.073 | 33.89 | 1.702 | 37.40 | 1.872 | 35.59 | 1.944 | 34.04 |
| | MASt3R | 1.158 | **49.65** | 0.687 | 41.41 | 1.052 | 41.26 | 1.108 | 41.72 | 1.180 | **50.15** |
| | VGGT | 0.722 | 36.95 | 1.393 | 45.30 | 1.647 | **46.67** | 1.637 | **46.59** | 2.172 | 44.84 |
| | COLMAP$^{LR}$ | 0.573 | 25.47 | 0.443 | 35.88 | **0.453** | 18.72 | 0.592 | 7.420 | 0.648 | 9.830 |
| | COLMAP$^{HR}$ | **0.114** | 36.57 | **0.149** | **46.13** | 0.953 | 30.25 | **0.169** | 25.63 | **0.357** | 17.47 |

### *4.2 Accuracy of Camera Poses*

Qualitative results in **Figure 5** demonstrate that the classic method produces the most accurate outcomes on large, high overlap datasets: the estimated camera positions and orientations show the smallest deviation from the ground truth poses in terms of spatial alignment and orientation consistency. In addition, VGGT demonstrates visually acceptable performance, with a higher proportion of estimated poses closely matching the ground truth. VGGT also reconstructs 100% of poses, whereas COLMAP$^{HR}$ achieves this in only 67% of cases. DUSt3R and MASt3R face challenges, with the global alignment process resulting in approximately 20% of the estimated poses deviating significantly from the ground truth, with some discrepancies exceeding several hundred meters.

Based on all evaluated cases, COLMAP$^{HR}$ achieves better camera pose center positions in all cases, as shown in **Table 5**. Note that single-image cases are excluded, as pose comparison is not meaningful due to perfect alignment. Interestingly, DUSt3R, MASt3R, and VGGT achieve superior orientation estimation in 75% of the evaluated cases, likely due to their learning-based methods, which leverage global scene context and robust feature matching to better handle orientation estimation.

It is also notable that many estimated poses exhibit large deviations from the ground truth, with errors reaching hundreds of meters or degrees. This prompts the question of how the results change when considering only inlier data points that meet established quality thresholds. An empirical threshold of 10 degrees for orientation error and 1 m for position error was applied to distinguish inliers from outliers, in line with thresholds commonly used in 3D reconstruction benchmarks (Sattler et al. 2018). Updated values after outlier filtering are shown in parentheses in **Table 5**. Red backgrounds indicate settings with at least one valid data point. The absence of parentheses denotes either no valid data (white background) or that all data points were valid and results are unchanged (red background). DUSt3R, MASt3R, and VGGT produce meaningful results primarily in scenarios with 2 or 5 input images, successfully reconstructing all poses and frequently generating a sufficient number of accurate estimates, although large errors occasionally occur. The limitations of DUSt3R and MASt3R stem from their pairwise matching and localization strategy, which is prone to cumulative errors as the number of input images increases. VGGT directly predicts point and depth maps with reasonable accuracy, but there remains significant potential for improvement, particularly by incorporating traditional strategies such as bundle adjustment. As expected, COLMAP fails in extremely small datasets due to fundamental limitations of the traditional SfM and MVS pipelines, which require sufficient image overlap and redundancy. Conversely, COLMAP provides accurate camera poses predominantly with larger datasets, achieving orientation errors below 24 degrees and position errors within 0.8 m.

In the low-overlap reconstruction experiment using 38 images, with or without thresholds applied (**Table 6**), COLMAP$^{HR}$ demonstrates a clear advantage in camera pose estimation across all scenarios, consistently achieving higher accuracy in both camera center localization and orientation. Even with minimal overlap, COLMAP$^{HR}$ maintains high accuracy, with position errors below 3 m and angular errors under 21 degrees. VGGT also produces accurate pose estimates in high-overlap cases, with center differences within 4 m. Additionally, it generates poses that meet the threshold requirements and can be identified as inliers, whereas all poses from DUSt3R and MASt3R are too scattered to qualify as inliers. MASt3R exhibits substantially larger errors, with position deviations exceeding 100 m and angular errors greater than 48 degrees. Overall, $\text{COLMAP}^{\text{HR}}$ provides substantial improvements, reducing camera center error by up to 99.77% and orientation error by up to 94.59%.

With thresholding applied, $\text{COLMAP}^{\text{HR}}$ achieves reconstruction success rates from 11% to 64% (**Table 7**). Considering the learning-based methods, only VGGT produces a limited number of valid poses for comparison, while the other methods do not yield any valid poses. Even under minimal overlap conditions, $\text{COLMAP}^{\text{HR}}$ successfully reconstructs a subset of images with acceptable accuracy, maintaining position errors below 0.7 m and angular errors under 10 degrees. However, despite the high accuracy of $\text{COLMAP}^{\text{HR}}$ reconstructed poses, the number of successfully reconstructed images is significantly limited. When the overlap rate falls below 40%, which is lower than COLMAP's typical operational range, $\text{COLMAP}^{\text{LR}}$ fails to reconstruct any valid poses within the defined thresholds and $\text{COLMAP}^{\text{HR}}$ reconstructs 51% of poses under these low-overlap conditions. The limitation results from a combination of low overlap and a relatively small image set of only 38 images, which is unusual for photogrammetry applications that generally use larger datasets.

In contrast, DUSt3R, MASt3R, and VGGT recover all camera poses even at 10% overlap, but DUSt3R and MASt3R produce significant errors, with position deviations exceeding 100 m and angular errors over 48 degrees, yielding no valid estimates after thresholding. VGGT generates comparatively better pose estimates, maintaining some valid results after thresholding, though still falling short of COLMAP's performance. These methods infer 3D structures and estimate camera parameters without requiring prior information about camera calibration or poses, offering greater flexibility but also introducing higher uncertainty in their performance. In real-world scenarios where ground truth is unavailable, VGGT offers an advantage by consistently providing pose estimates even when COLMAP fails. These estimates can serve as initial guesses and be further refined using traditional photogrammetric techniques such as bundle adjustment.

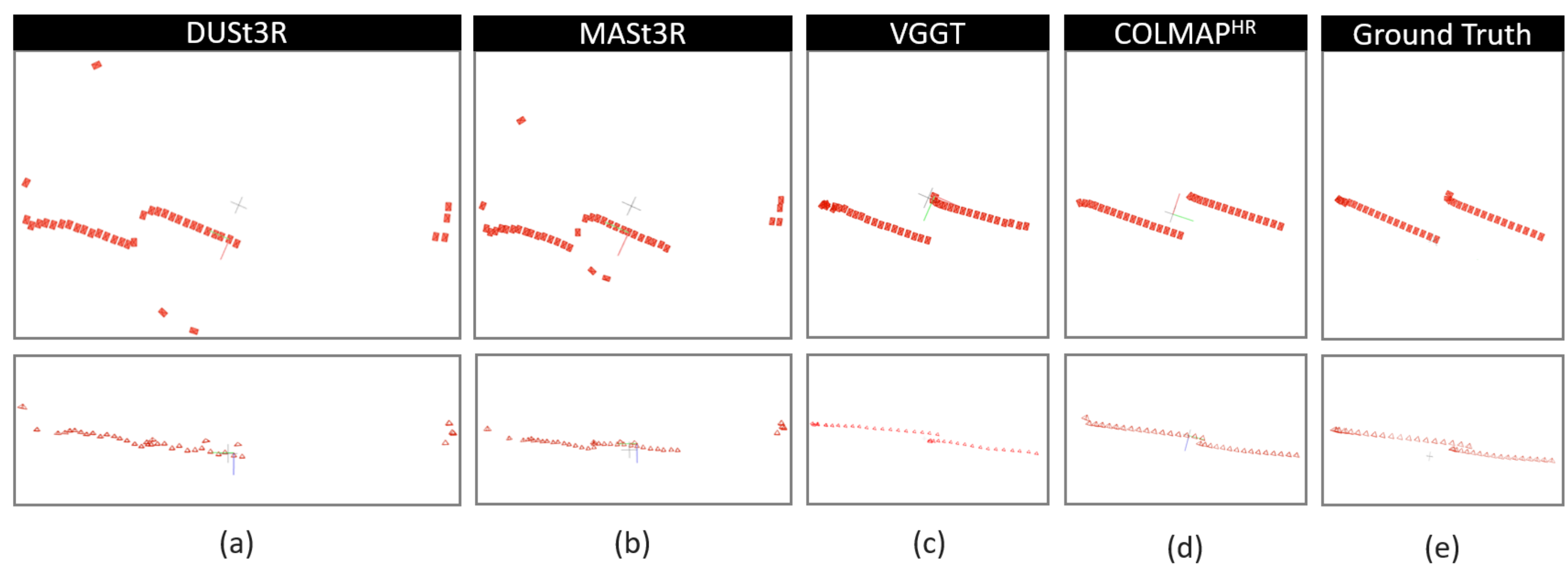

**Figure 5**. Estimated camera poses from 38 images in Dataset-3. (a), (b), (c), and (d) show the estimated camera poses from DUSt3R, MASt3R, VGGT, and COLMAP$^{HR}$, respectively; (e) shows the ground truth camera poses. The first row presents top-down views, and the second row shows front views.

**Table 5.** Quantitative evaluation of camera pose estimation across three datasets using 2, 5, 10, and 38 images. Cen. D. (center distance) represents the distance between the estimated camera center and the ground truth. Ang. D. (angle difference) measures the orientation error. Succ. R. (success rate) indicates the percentage of successfully reconstructed camera poses relative to the total number of poses. For each method and overlap, the main value uses all reconstructed poses. Values in parentheses use only inliers (center distance < 1 m and angle difference < 10°). If no parentheses appear, no inliers were found.

| **Dataset** | **Method** | **2 Image** | | **5 Images** | | | **10 Images** | | | **38 Images** | | |
|---|---|---|---|---|---|---|---|---|---|---|---|---|
| | | **Ang. D. (°)** | **Succ. R.(%)** | **Cen. D. (m)** | **Ang. D. (°)** | **Succ. R.(%)** | **Cen. D. (m)** | **Ang. D. (°)** | **Succ. R.(%)** | **Cen. D. (m)** | **Ang. D. (°)** | **Succ. R.(%)** |
| Dataset-1 | DUSt3R | **4.390 (4.390)** | **100 (100)** | 3.688 | 36.76 | **100** | 4.149 | **5.695** | **100** | 63.68 | 16.41 | **100** |
| | MASt3R | 24.24 | **100** | 6.772 | **1.216** | **100** | 34.86 | 41.42 | **100** | 62.14 | 8.220 | **100** |
| | VGGT | 47.68 | **100** | 0.432 | 19.54 | **100** | 0.390 | 19.74 | **100** | 2.803 | 19.87 | **100** |
| | COLMAP$^{LR}$ | - | 0 | 0.462 | 76.56 | **100** | 0.862 | 14.79 | **100** | 1.204 (0.537) | 14.45 (8.274) | **100** (21) |
| | COLMAP$^{HR}$ | - | 0 | **0.115** | 15.15 | **100** | **0.160** (**0.159**) | 10.49 (**9.293**) | **100** (**80**) | **0.113** | **2.506** | **100** |
| Dataset-2 | DUSt3R | **0.837 (0.837)** | **100 (100)** | 0.377 (0.377) | **1.687 (1.687)** | **100 (100)** | 0.813 (0.578) | **3.325** (**3.285**) | **100** (60) | 66.24 | **2.108** | **100** |
| | MASt3R | 1.738 (1.738) | **100 (100)** | 11.251 | 5.041 | **100** | 58.82 | 49.73 | **100** | 180.3 | 56.18 | **100** |
| | VGGT | 17.11 | **100** | 0.391 | 6.435 | **100** | 0.657 (**0.494**) | 3.977 (3.925) | **100** (80) | 4.582 | 19.64 | **100** |
| | COLMAP$^{LR}$ | - | 0 | 0.368 | 48.45 | **100** | 0.702 (0.625) | 6.730 (6.667) | **100** (**90**) | 0.894 (0.686) | 11.51 (4.204) | **100** (34) |
| | COLMAP$^{HR}$ | 70.02 | **100** | **0.120** | 50.42 | **100** | **0.190** | 24.70 | **100** | **0.196 (0.196)** | 6.212 (6.212) | **100 (100)** |
| Dataset-3 | DUSt3R | 69.49 | **100** | 7.560 | 64.361 | **100** | 30.83 | 76.14 | **100** | 104.8 | 33.50 | **100** |
| | MASt3R | 28.56 | **100** | 4.362 | **8.408** | **100** | 94.78 | 36.12 | **100** | 122.6 | 69.60 | **100** |
| | VGGT | **26.32** | **100** | 0.499 | 102.8 | **100** | 0.573 | 120.6 | **100** | 3.318 | 21.10 | **100** |
| | COLMAP$^{LR}$ | - | 0 | - | - | 0 | - | - | 0 | **0.724** | 17.89 | 0 |
| | COLMAP$^{HR}$ | - | 0 | **0.089** | 30.03 | 80 | **0.180** | **18.32** | 90 | 0.823 (**0.475**) | **14.19** (**9.595**) | 90 **(47)** |

**Table 6.** Quantitative evaluation of camera pose estimation across three datasets with varying image overlaps. Cen. D. (center distance) represents the distance between the estimated camera center and the ground truth. Ang. D. (angle difference) measures the orientation error. For each method and overlap, the main value uses all reconstructed poses. Values in parentheses use only inliers (center distance < 1 m and angle difference < 10°). If no parentheses appear, no inliers were found.

| **Dataset** | **Method** | **Overlap: 70%** | | **Overlap: 55%** | | **Overlap: 40%** | | **Overlap: 25%** | | **Overlap: 10%** | |
|---|---|---|---|---|---|---|---|---|---|---|---|
| | | **Cen. D. (m)** | **Ang. D. (°)** | **Cen. D. (m)** | **Ang. D. (°)** | **Cen. D. (m)** | **Ang. D. (°)** | **Cen. D. (m)** | **Ang. D. (°)** | **Cen. D. (m)** | **Ang. D. (°)** |
| Dataset-1 | DUSt3R | 47.21 | 141.3 | 61.15 | 92.15 | 74.63 | 78.91 | 97.64 | 47.41 | 111.5 | 48.11 |
| | MASt3R | 50.57 | 58.08 | 58.79 | 19.78 | 66.03 | 172.2 | 92.46 | 42.63 | 109.4 | 47.37 |
| | VGGT | 1.534 (0.683) | **92.07** (7.790) | 1.600 (0.742) | 92.95 (7.033) | **2.532** | 90.87 | **4.427** | 88.23 | 7.639 | 86.92 |
| | COLMAP$^{LR}$ | 1.221 | **8.097** | 2.675 | 15.16 | 10.91 | **41.10** | 41.10 | 37.59 | - | - |
| | | (0.487) | (8.203) | (0.909) | (**2.403**) | (0.844) | (**0.225**) | | | | |
| | COLMAP$^{HR}$ | **0.152** **(0.125)** | 17.62 (**1.946**) | **0.182** **(0.194)** | **8.609** (8.244) | 2.907 (**0.378**) | 100.2 (7.572) | 30.25 **(0.661)** | **29.96** **(9.928)** | **1.140** **(0.607)** | **15.49** **(9.278)** |
| Dataset-2 | DUSt3R | 60.17 | 120.7 | 87.84 | 166.5 | 121.0 | 58.79 | 155.7 | 116.6 | 143.4 | 38.89 |
| | MASt3R | 60.81 | 113.8 | 96.72 | 92.30 | 144.3 | 141.0 | 157.0 | 155.6 | 149.3 | 130.1 |
| | VGGT | 1.886 (0.750) | **6.905** (3.556) | 3.592 (0.863) | **8.920** (4.413) | 86.15 | 20.93 | 48.45 | 18.25 | 74.07 | 83.29 |
| | COLMAP$^{LR}$ | 0.938 (0.576) | 7.249 **(1.280)** | 1.295 (0.749) | 9.014 **(0.184)** | 23.60 | 30.33 | - | - | - | - |
| | COLMAP$^{HR}$ | **0.140** (**0.114**) | 7.966 (2.025) | **0.266** **(0.268)** | 10.54 (2.076) | **0.450** **(0.456)** | **7.527** **(1.506)** | **1.043** **(0.565)** | **7.627** **(7.003)** | **0.734** **(0.475)** | **12.21** **(8.581)** |
| Dataset-3 | DUSt3R | 56.95 | 128.4 | 90.43 | 140.5 | 105.4 | 45.53 | 100.0 | 172.7 | 101.2 | 161.4 |
| | MASt3R | 59.08 | 175.3 | 89.33 | 157.2 | 106.1 | 95.84 | 101.1 | 101.2 | 101.1 | 166.8 |
| | VGGT | 1.896 (0.631) | **8.025** (3.936) | 4.497 (0.706) | 7.924 (8.438) | 4.750 | 143.4 | 5.708 | 96.48 | 6.391 | 91.06 |
| | COLMAP$^{LR}$ | 1.594 (0.631) | 29.27 (1.895) | 2.146 (0.776) | 15.16 (3.434) | 3.884 (0.999) | 14.49 (9.183) | 62.90 | 82.23 | 97.55 | 123.4 |
| | COLMAP$^{HR}$ | **0.219** **(0.160)** | 14.90 **(1.728)** | **0.373** **(0.349)** | **9.340** **(2.396)** | **1.022** **(0.538)** | **8.770** **(6.465)** | **3.335** **(0.653)** | **9.356** **(9.136)** | **2.737** **(0.500)** | **20.65** **(3.906)** |

**Table 7.** Success rate (%) of reconstructed images across different overlap levels. The success rate is computed as the number of successfully reconstructed images divided by the total number of images.

| **Method** | **Success Rate at Different Overlap Levels (%)** | | | | |
|---|---|---|---|---|---|
| | **70%** | **55%** | **40%** | **25%** | **10%** |
| DUSt3R | **100** (0) | **100** (0) | **100** (0) | **100** (0) | **100** (0) |
| MASt3R | **100** (0) | **100** (0) | **100** (0) | **100** (0) | **100** (0) |
| VGGT | **100** (10) | **100** (6) | **100** (0) | **100** (0) | **100** (0) |
| COLMAP$^{LR}$ | 75 (27) | 84 (11) | 60 (2) | 20 (0) | 13 (0) |
| COLMAP$^{HR}$ | 85 (**64**) | 61 (**53**) | 85 (**35**) | 85 (**22**) | 51 (**11**) |

### *4.3 Scalability Evaluation*

All four methods were evaluated on the standard 38-image dataset, but only VGGT and COLMAP can process larger image sets. Therefore, we conducted an additional scalability experiment with 191 images.

Visualization results for Dataset-2 are presented in **Figure 6**. The VGGT reconstructions exhibit pronounced inconsistencies in point cloud alignment, such as overlapping buildings, repeated occurrences of the same structures at multiple locations, and road segments that are interpolated in ways inconsistent with the actual scene geometry. In comparison, COLMAP generates three separate models, but each reconstructed point cloud is internally consistent and does not display significant misalignment. **Table 8** presents the quantitative results for dense point cloud and camera pose accuracy. VGGT demonstrates higher point cloud errors, reaching up to 6 m, which represents approximately an 85% increase compared to COLMAP$^{HR}$'s. Additionally, camera pose estimates produced by VGGT may exhibit drift of up to 42 m. Substantial errors in both point cloud and camera pose estimation mean VGGT cannot yet deliver reliable or usable previews for the areas of interest, and it is still not suitable as a standalone solution for large-scale aerial photogrammetry, although VGGT demonstrates better scalability than the other end-to-end approaches.

### *4.4 Computation Time*

DUSt3R/MASt3R are significantly faster than COLMAP, and VGGT can be remarkably faster than DUSt3R/MASt3R as well. For instance, in the 38-image case (Table 9), MASt3R requires only 9% of COLMAPHR's processing time, while VGGT operates at just 12% of MASt3R's processing time, making VGGT particularly suitable for compute-constrained environments. The substantial reduction in processing time is likely due to VGGT's multi-image training paradigm, which enables the network to natively perform multiview triangulation. In contrast, DUSt3R relies on separate pairwise triangulations that are later averaged, resulting in less efficient alignment procedures.

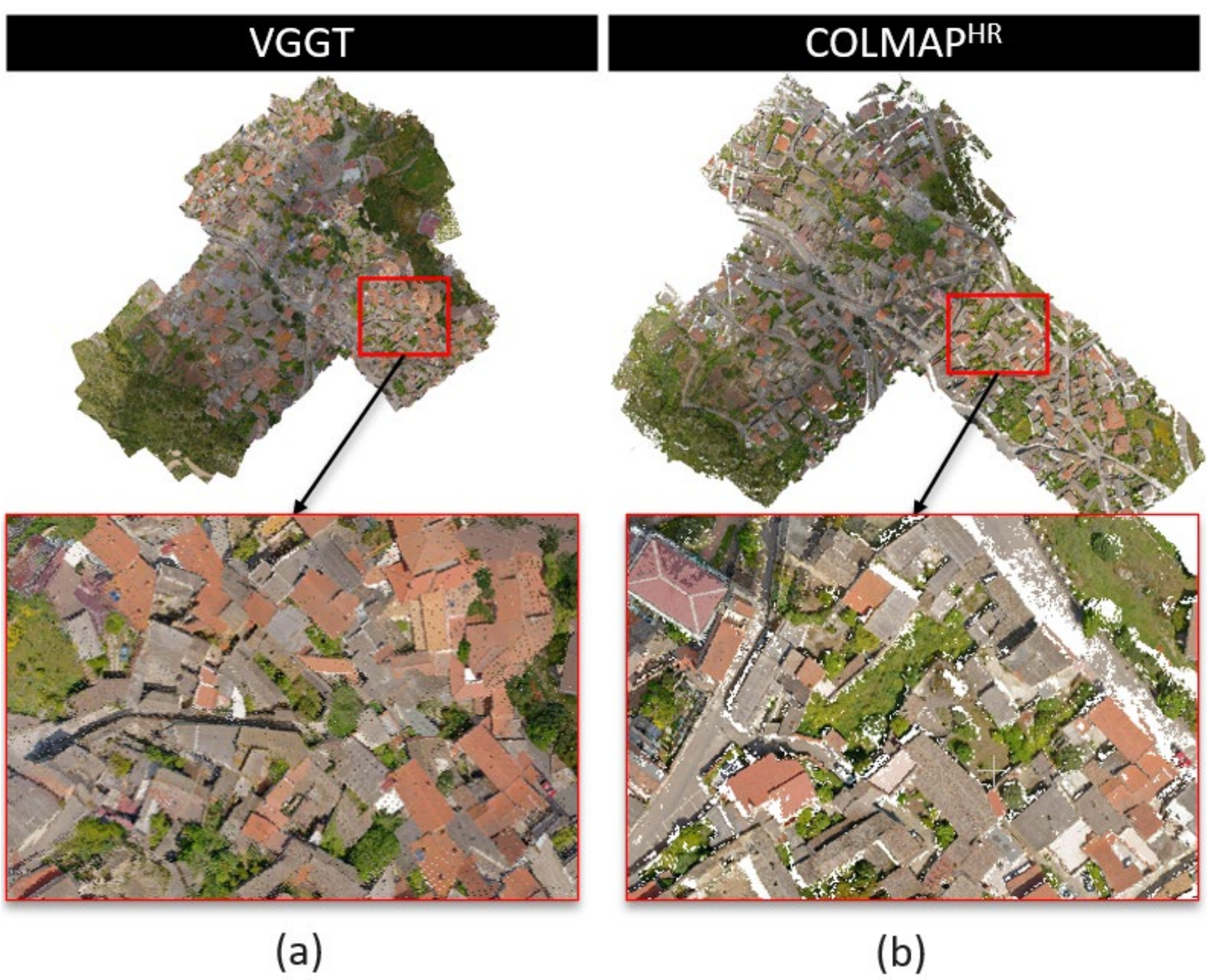

**Figure 6**. Reconstruction models for 191-image experiment on Dataset-2: (a) VGGT, (b) COLMAP$^{HR}$.

**Table 8.** Point cloud and camera pose evaluation of VGGT and COLMAP$^{HR}$ on three benchmark datasets. For camera poses, the values in parentheses are for inliers (center distance <1 m, angle difference <10°). Red cells indicate at least one valid inlier; white cells mean no inliers were found for that setting.

| **Dataset** | **Method** | **Point clouds** | | **Camera poses** | | |
|---|---|---|---|---|---|---|
| | | **Accu. (m)** | **Comp. (%)** | **Cen. D. (m)** | **Ang. D. (°)** | **Succ. R. (%)** |
| **Dataset-1** | VGGT | 2.936 | 35.44 | 10.41 | 101.6 | **100** |
| | COLMAP$^{HR}$ | **0.123** | **75.06** | **0.524 (0.352)** | **34.501 (7.192)** | 96 (**69**) |
| **Dataset-2** | VGGT | 5.991 | **45.40** | 42.22 | 81.97 | **100** |
| | COLMAP$^{HR}$ | **0.876** | 42.77 | **0.765 (0.4551)** | **13.84 (5.352)** | 96 (**48**) |
| **Dataset-3** | VGGT | 2.988 | 38.83 | 31.48 | 80.82 | **100** |
| | COLMAP$^{HR}$ | **0.197** | **64.70** | **0.526 (0.351)** | **15.01 (9.898)** | 94 (**75**) |

**Table 9.** Average processing time (in seconds) for image sets of varying sizes across different methods.

| Method | Time Cost (s) | | | | | |
|---|---|---|---|---|---|---|
| | **1** | **2** | **5** | **10** | **38** | **191** |
| DUSt3R | **9** | **9** | 11 | 20 | 191 | - |
| MASt3R | **9** | **9** | 12 | 22 | 208 | - |
| VGGT | **9** | **9** | **10** | **12** | **24** | **103** |
| COLMAP$^{LR}$ | - | - | 41 | 87 | 370 | - |
| COLMAP$^{HR}$ | - | - | 271 | 568 | 2349 | 5280 |

*4.5 A case study in mesh completion of regions with low image overlap*

We apply learning-based methods as mesh completion step to repair missing parts in an imperfect 3D model. As an example, **Figure 7** (left) shows a model reconstructed from UAV images in our self-collected dataset with nadir and near oblique views. The dataset is described in Section 3. In the COLMAP reconstruction, coverage is sparse near the edge of the area of interest, and a tall building has an incomplete facade. This pattern is common with fixed flight plans over complex sites, where low and simple areas receive many images while tall and complex structures receive few. In the baseline COLMAP run, the facade fails to reconstruct because nearby views do not form tracks long enough for triangulation, even though one view contains rich facade texture. With a learning-based method we produce a single image 3D point map for that view, align it to the scene, and fuse it with the model, as shown in **Figure 7** (right). This recovers facade structure, reduces the gap, and improves completeness.

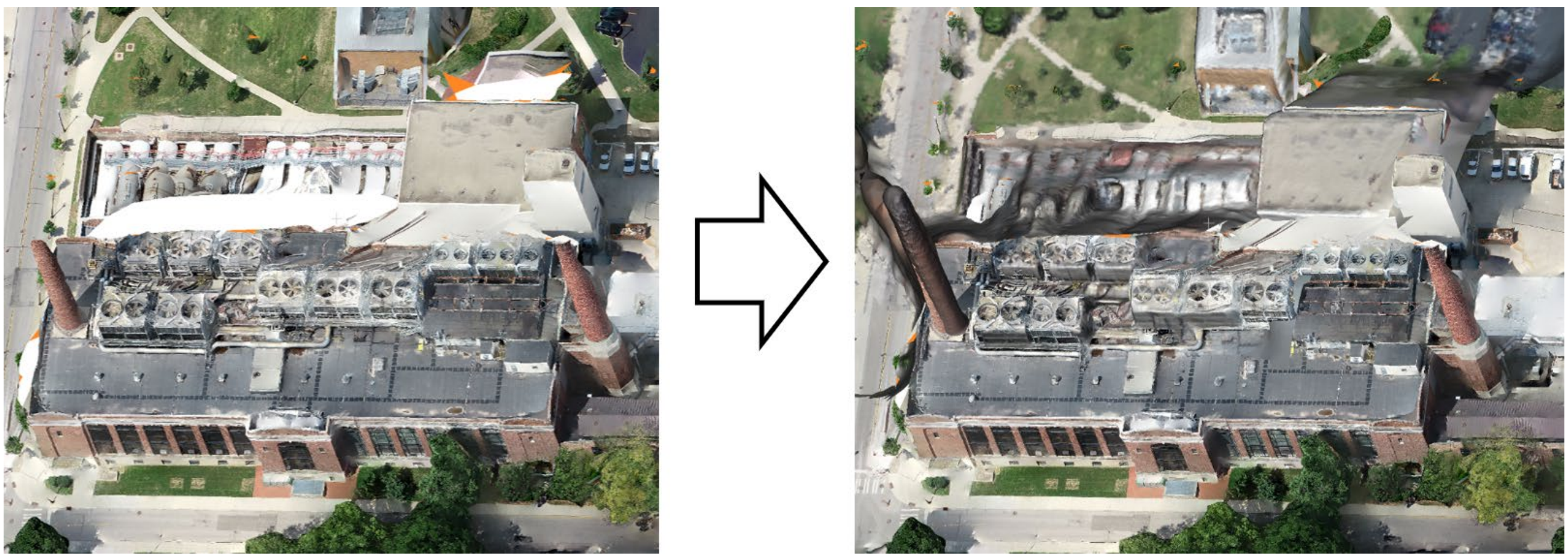

**Figure 7** Filling a facade gap with a learned monocular point map. Left COLMAP reconstruction with a missing facade due to sparse overlap. Right the model after adding a single image 3D point map from DUSt3R, aligned and fused with the COLMAP result, which closes the gap and improves completeness.

Learning-based methods work well in many practical cases. They are especially helpful in mission critical settings where recollection is impossible, and overlap is weak. In multi date aerial collections, seasonal and radiometric changes reduce classical keypoint matches and limit triangulation, and deep models have been proposed to handle illumination and appearance differences (Huang et al. 2024; Albanwan and Qin 2022). Moreover, prior work in satellite settings reports that deep models can yield better results on specific regions such as flat ground (Sadeq 2025). In one-time collections, many frames may be missing, and view angles can vary widely. There are also studies that apply learning-based methods directly to heritage 3D reconstruction (Ge et al. 2024; Mazzacca et al. 2023). Learned methods also enable single image 3D point maps that can be aligned to a scene, which gives a practical way to add structure in places where pairwise matching is unreliable. Therefore, these results justify the use of learning-based matching for sparse coverage, multi date collections, and historical archives where traditional pipelines struggle.

## 5 Conclusion

This study critically assesses state-of-the-art learning-based direct 3D reconstruction methods (DUSt3R, MASt3R, and VGGT) against the classic COLMAP pipeline on the UseGeo photogrammetry dataset. We evaluate scenarios that reflect both typical and challenging conditions in aerial photogrammetry, with input image counts from 1 to 191 and overlap levels from approximately 10% to 70%. Unlike general computer vision datasets, aerial photogrammetry involves large-scale outdoor scenes, highly regular acquisition geometry, and industry-standard requirements for geometric accuracy and completeness.

VGGT and MASt3R perform impressively in scenarios characterized by minimal image counts or low overlap, producing dense point clouds with accuracy up to 0.4 m and completeness as high as +50%, substantially outperforming COLMAP, which either fails or yields extremely poor results (as low as 8% completeness). However, COLMAP performs best in standard photogrammetric scenarios involving larger image sets and higher overlaps, with errors as low as 0.06 m (compared to more than 1 m for VGGT) and completeness of up to 74% (versus 36% for DUSt3R). For camera pose estimation, COLMAP surpasses others in nearly all standard scenarios, with the exception of cases involving only two input images. However, VGGT‘s advantage is its ability to recover image poses where other methods fail.

Among these learning-based solutions, VGGT uniquely extends processing capability from dozens to hundreds of images and can produce camera poses that meet inlier criteria. Nevertheless, VGGT cannot serve as a replacement for traditional SfM and MVS pipelines in typical photogrammetric applications, as its strongest performance is restricted to narrow cases, mainly one or two images, and its flexibility is lower than COLMAP. Instead, VGGT is most valuable as a supplement, for example to fill model gaps or to recover initial poses in sparse-image situations. Although VGGT achieves significant time savings, requiring only 1% of COLMAP's processing time for the 38-image case, its scalability is currently limited to the hundreds range.

Overall, our findings indicate that COLMAP remains the most robust and versatile solution for aerial photogrammetry datasets, particularly in standard, high-overlap scenarios. Nevertheless, VGGT exhibits distinct advantages when inputs are extremely limited and when computational efficiency is a priority. These attributes position VGGT as a promising supplementary approach for challenging or resource-constrained photogrammetric applications.

To enhance VGGT's accuracy and its capacity to process higher-resolution imagery, several strategic improvements are recommended. Although VGGT currently exhibits limitations in camera pose accuracy, its estimated poses can serve as effective initial approximations that enable further refinement through traditional SfM and MVS pipelines, such as by applying subsequent bundle adjustment. In addition, fine-tuning on specialized aerial or aerial-ground datasets, such as AerialMegaDepth, may significantly boost performance. Together, these improvements are expected to collectively strengthen the robustness, accuracy, and practical applicability of VGGT in photogrammetric workflows.

## Funding

This research was partially supported by the Intelligence Advanced Research Projects Activity (IARPA) via Department of Interior/ Interior Business Center (DOI/IBC) contract number 140D0423C0075. The U.S. Government is authorized to reproduce and distribute reprints for Governmental purposes notwithstanding any copyright annotation thereon. Disclaimer: The views and conclusions contained herein are those of the authors and should not be interpreted as necessarily representing the official policies or endorsements, either expressed or implied, of IARPA, DOI/IBC, or the U.S. Government. It is also supported by the Office of Naval Research (Award No. N000142012141 and N000142312670).

## Notes on contributor(s)

***Xinyi Wu*** received the B.S. degree in Electrical Engineering and Its Automation from North China Electric Power University, China, in 2020, and the M.S. degree in Electrical and Computer Engineering from The Ohio State University, Columbus, OH, USA, in 2025. She is currently a Ph.D. student in Civil, Environmental and Geodetic Engineering at The Ohio State University, with research interests in 3D reconstruction and computer vision.

***Steven Landgraf*** received the B.Sc. and M.Sc. degrees in Geodesy and Geoinformatics from Karlsruhe Institute of Technology, Germany, in 2018 and 2020, respectively, and completed his Dr.-Ing. degree from Karlsruhe Institute of Technology in 2025. He is currently a Postdoctoral Researcher in the Machine Vision Metrology group at the Institute of Photogrammetry and Remote Sensing, Karlsruhe Institute of Technology. His research interests include machine learning, photogrammetry, and explainable AI.

***Markus Ulrich*** is a distinguished academic and industry expert with over twenty years of experience bridging the realms of academia and industry. Currently a Professor for Machine Vision Metrology at the Institute of Photogrammetry and Remote Sensing at the Karlsruhe Institute of Technology (KIT), his journey began with a PhD degree at the Department of Civil, Geo, and Environmental Engineering of Technical University of Munich (TUM). Prior to academia, he served as the Invention and Patent Manager at MVTec Software GmbH, Munich, and headed the research team at the same institution. In 2017, he completed his habilitation (venia legendi) and was appointed as a Privatdozent (Lecturer) at KIT. His research interests include machine vision, close-range photogrammetry, machine learning, and their applications in industry for quality inspection, automation, and robotics.

***Rongjun Qin*** (Senior Member, IEEE) is a Full Professor of the Department of Civil, Environmental and Geodetic Engineering, and the Department of Electrical and Computer Engineering, The Ohio State University, Columbus, OH, USA. He received the B.S. degree in computational mathematics and the M.S. degree in photogrammetry and remote sensing from Wuhan University, Wuhan, China, in 2009 and 2011, respectively, and the Ph.D. degree in photogrammetry and remote sensing from ETH Zürich, Zürich, Switzerland, in 2015., His research interests include photogrammetric 3-D reconstruction, remote sensing image classification, UAV image processing, image dense matching, and change detection. His research seeks computational solutions to various geometric and interpretation problems in an urban context using imaging sensors, such as aerial/UAV imagery, LiDAR, and satellite multispectral/hyperspectral images. He is the author of the RPC stereo processor (RSP) and multi-stereo processor (MSP) used for reconstructing 3-D information from 2-D images with high quality. Prof. Qin's awards include the First Prize in the Mathematical Modeling Contest and several other prominent scholarship awards. He is also chairing the working group "Satellite Constellation for Remote Sensing" of the International Society for Photogrammetry and Remote Sensing Commission.

**ORCID**

Xinyi Wu https://orcid.org/0009-0004-4437-7157

Steven Landgraf https://orcid.org/0000-0002-2056-567X

Markus Ulrich https://orcid.org/0000-0001-8457-5554

Rongjun Qin https://orcid.org/0000-0002-5896-1379

**Disclosure statement**

No potential conflict of interest was reported by the author(s).

**Data availability statement**

The original data presented in the study are openly available in the UseGeo at https://usegeo.fbk.eu/.

**Author Contributions**

Conceptualization, X.W. and R.Q..; methodology, X.W. and S.L and R.Q.; validation, X.W.; formal analysis, X.W.; investigation, X.W. and S.L.; resources, M.U. and R.Q.; data curation, X.W. and S.L.; writing—original draft preparation, X.W. and S.L; writing—review and editing, M.U. and R.Q.; visualization, R.Q.; supervision, R.Q.; project administration, R.Q.; funding acquisition, R.Q.